\documentclass[conference]{IEEEtran}
\IEEEoverridecommandlockouts
\usepackage{cite}
\usepackage{amsmath,amssymb,amsfonts}
\usepackage{algorithmic}
\usepackage{graphicx}
\usepackage{textcomp}
\usepackage{cite}
\usepackage{xcolor}
\usepackage{amssymb}
\usepackage{amsmath}
\usepackage{booktabs}
\def\BibTeX{{\rm B\kern-.05em{\sc i\kern-.025em b}\kern-.08em
    T\kern-.1667em\lower.7ex\hbox{E}\kern-.125emX}}

\begin{document}

\title{
Grounding Implicit Goal Description for Robot Indoor Navigation Via Recursive Belief Update
}

\author{Rui Chen, Jinxin Zhao, Liangjun Zhang
\thanks{Baidu Research Institute, Baidu USA, 1195 Bordeaux Drive, Sunnyvale, California, 94089 {\tt\small \{ruichen, jinxinzhao, liangjunzhang\}@baidu.com}}

\\
}

\maketitle 

\begin{abstract}
Natural language-based robotic navigation remains a challenging problem due to the human knowledge of navigation constraints, and destination is not directly compatible with the robot knowledge base. In this paper, we aim to translate natural destination commands into high-level robot navigation plans given a map of interest. We identify grammatically associated segments of destination description and recursively apply each of them to update a belief distribution of an area over the given map. We train a destination grounding model using a dataset of single-step belief update for precise, proximity, and directional modifier types. We demonstrate our method on real-world navigation task in an office consisting of 80 areas. Offline experimental results show that our method can directly extract goal destination from unheard, long, and composite text commands asked by humans. This enables users to specify their destination goals for the robot in general and natural form. Hardware experiment results also show that the designed model brings much convenience for specifying a navigation goal to a service robot.
\end{abstract}


\section{Introduction}

\begin{figure}[ht!]
    \centering
    \includegraphics[width=\columnwidth]{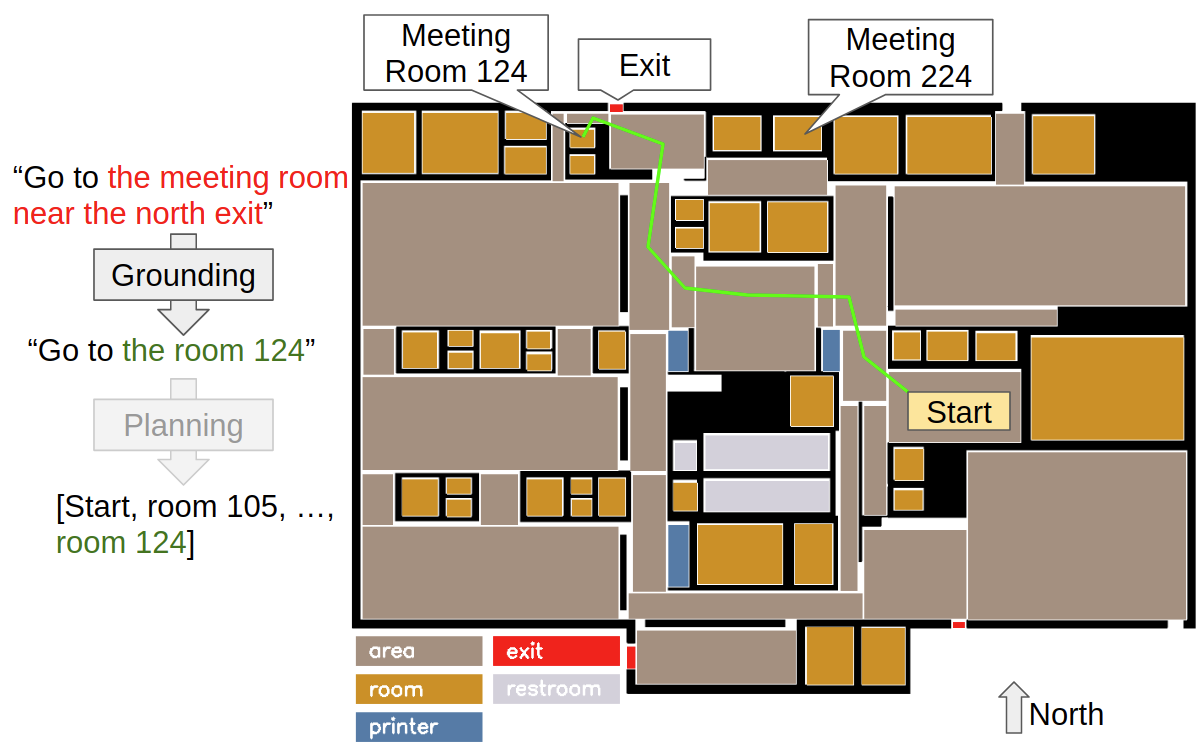}
     \caption{Destination grounding translates implicit destination descriptions (red) into robot-compatible locations (green).
    }
    \label{fig:intro}
\end{figure}

In human-robot interaction, natural language is one of the most desirable forms of communication between users and robots \cite{Tellex11,Jia18}. However, the interpretation of natural language remains an extremely hard problem for robots. One major issue is that even with the successful conversion of speech to text, there is still a considerable gap between text and its appropriate interpretation. One scenario where this issue exists is language-based robot navigation. Consider the map layout of an office in Fig.~\ref{fig:intro}. If one wants the robot to deliver a document to meeting room 124, he/she would command differently depending on his/her knowledge about the office. If the user knows the exact room number ``124'', he/she would say ``Go to room 124''. Otherwise, the user might refer to the meeting room with respect to a nearby location, an alternative but very intuitive command would be ``Go to the meeting room near the north exit.'' Despite the tiny modification of the original command, it has already become non-trivial for robots to understand due to the reasoning required.
The above example represents a common yet challenging situation in natural language-based navigation where the users' knowledge about desired destinations is not directly compatible with the robot knowledge base. This manifests especially when the user is unable to uniquely refer to the destination without proper explanation, such as ``the meeting room near the north exit.'' This destination can be easily made unique by assigning a coordinate to it, which, unfortunately, is normally done by robots only. A common workaround for human users is to describe by reference to landmarks or places easier to describe, yielding the alternative command mentioned previously. The major difficulty here is that although it is feasible for robots to memorize a map with extremely high fidelity, it is impractical to store all possible relations and interactions of map locations. However, the latter one is more commonly invoked in human language; in this way, people only need to memorize a few map elements and can potentially indicate anywhere in the map by adding adequate references and implications. In this paper, we formally define the above problem as \textit{implicit destination} in navigation instructions and aim at a solution to enabling robots to correctly ground implicit destination descriptions to specific locations on a map of interest.

Recently, interesting results for translating  navigation  instructions to  a  high-level plan are discussed in \cite{chen2011learning,zang2018translating}, where a graph is first built according to the preliminary knowledge of the environment. The detailed language text commands then guide the graph search algorithms to obtain the viable edges, which eventually form the navigation plan. However, this approach requires rather rich preliminary information, thus knowing the destination is enough to generate path plan. Hence the text commands appear to be redundant. In literature, a general frame of reinforcement learning (RL) has been investigated for the purpose of grounding the natural languages into robot behaviors \cite{Tellex18,gopalan2018sequence,oh2019planning}. Normally the system description leverages a system model related to Markov Decision Process (MDP) such as object-orientated MPD (OO-MDP) or
linear temporal logic (LTL). Based on such robot behavior model, RL algorithms are explored in order to build a mapping from natural language to certain action. Nonetheless, applications of such framework is restricted, due to the limited number of robotic tasks that can be transferred into a MDP model. Combining vision and language for navigation has also attract much research attention \cite{tan2019learning,thomason2019vision,huang2019transferable}. These approaches normally try to reveal the correspondence between the image and the text, based on which the next step action is selected among a finite set of pre-defined actions.

In this  paper,  we aim to translate  natural  destination commands  into high-level robot navigation plans given  a  map of  interest, where each area in the map is associated with as a tuple of strings with unique area id, area category, and area name. 
Instead of directly taking the whole destination noun phrase as input, we decompose it into a sequence of grammatically associated segments, namely - \textit{modifier} and recursively apply each of them to update the belief distribution based on the prior belief. We categorize modifiers into ``dummy'', ``proximity'', ``precise'', or ``directional'' according to how they update the prior. Each modifier type also implies constraints on the type of prior it applies to; ``precise'' believes refer to specific areas, while ``proximity '' believes refer to an orientation or proximity relation with respect to its prior. 

We demonstrate  our method  on  real-world  navigation  task  in  an  office  consisting of 80 areas. The robot's goal is to ground the destination description given by a user to the office map. Since the language involved in single updates are highly limited, we analytically generate training data based on a finite rule set for all different types of modifiers for a single belief update. We then train the learnable update functions on single update steps for each type of modifier by minimizing the total losses of all supervised terms that are applicable for each update type. Our model achieves ~$90\%$ accuracy for the area grounding of single-step precise belief update. We further demonstrate a composite belief update on realistic human instructions. Experimental results show that our method can directly extract  goal  destination from  unlabeled,  long  and  composite  text commands asked by humans. This enables users to specify their destination  goals  for  the  robot in  general  and  naturalistic  form.
\section{Problem Definition}
\label{sec:prob_form}

Our goal is to enable robots to understand highly naturalistic language commands containing 
implicit destinations, and generate high level path plans given a map of interest. We assume that the map $m\in\mathcal{M}$ is already segmented into a set of areas of interest $\{a_i\}_{i=1}^S$, each having boundary $B_i$ within the map boundary $B_0$. Each area $a_i$ is associated with as a tuple of strings that are commonly mentioned, such as unique area id, area category, and area name if applicable. We assume that at most $N$ words are assigned to each area of interest. We then define our high level path plan $\mathbf{x}\in\mathcal{X}$ as a sequence of areas, or $\mathbf{x}:=(a_{k_j})_{j=1}^{|\mathbf{x}|}$, where $k_j\in[1,S],\forall j$. On the other hand, each instruction $w\in\mathcal{W}$ is a command with implicitly described destination. Our goal is a function $\mathcal{F}: \mathcal{M}\times\mathcal{W}\rightarrow\mathcal{X}$ that interprets natural language navigation instructions to robot-compatible high level path plans. Since we allow implicit destinations, our goal can be decomposed into two sub-tasks: goal grounding and path planning. In the first sub-task, we identify the noun phrase in $w$ referring to the destination and ground it on the given map as an area $a^*$. In the second task, we generate a high level path plan $\mathbf{x}^*$ according to the grounded destination as well as the current robot location. In this paper, we primarily focus on the 
 destination grounding task and apply a graph search planning to fulfill the intermediate way points. In the next section, we present our implicit destination grounding algorithm.




\section{Implicit Goal Grounding}



Given a navigation instruction $w$, we first utilize state-of-the-art natural language processing (NLP) tools to parse its grammatical structure. Assuming that $w$ takes a simple form that only specifies the destination, such as ``go to the meeting room near the north exit'', we first perform semantic role labeling \cite{SemLabeling19} to extract the noun phrase that describes the destination. Then, our goal is to ground the destination, ``\textbf{meeting room near the north exit}'', to a specific area $a^*$ on the map that can be fed to downstream path planning modules. The key to achieving this goal is to view the noun phrase as constructing a chain of spatial relationships that points to the final destination. In the following sections, we provide a formal explanation and build a recursive belief update algorithm that utilizes the grammatical structure of destination noun phrases.

\subsection{Goal Grounding via Dependency Parsing}

Consider the scenario where the user wants to  locate a meeting room in Fig.~\ref{fig:intro}, but does not know its room id. Since only mentioning ``a meeting room'' would lead to ambiguity, assuming multiple meeting rooms exit, it is natural for the user to append a spatial constraint with respect to another location that is easier to uniquely identify. In this case, ``the north exit'' is a good choice, since there is only one exit to the north of the map. In practice, this command can be easily understood by other humans by first locating the north exit and then looking for the nearest meeting room. In order to enable the robot to also interpret this implicit goal description, we establish our method by resembling the later process, which takes advantage of the dependency structure of the phrase.

\begin{figure}[ht!]
    \centering
    \includegraphics[width=\columnwidth]{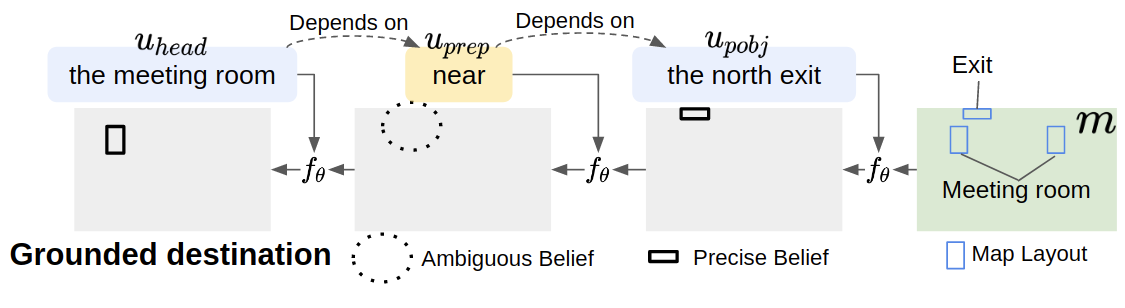}
     \caption{
     Given an composite goal description ``the meeting room near the north exit'', our destination grounding algorithm start from the dummy prior (green) and recursively performing general updates (see Eq. \eqref{eq:pobj} to \eqref{eq:h}). Finally, one of the two meeting rooms is located considering the spatial constraint introduced by the modifiers ``near the north exit''. Depending on the modifier semantic, intermediate belief $b_k$ can be precise (solid) or ambiguous (dashed). The map layout and area information is assumed known (blue).
    }
    \label{fig:belief_update}
\end{figure}

Given a simple noun phrase, we can again use state-of-the-art NLP annotation tools \cite{ManningICLR17} to parse that phrase into a dependency structure with head words $u_h$, prepositions $u_{prep}$, and objectives of preposition $u_{pobj}$. In navigation instructions, the preposition $u_{prep}$ normally specifies a spatial relationship between head word $u_h$ and the preposition objective $u_{pobj}$. In other words, $u_{pobj}$ serves as the dependency of $u_{h}$ and contributes to its unique grounding. Based on the above observation, we can first ground $u_{pobj}$ to the map $m$ and generate a belief of location reference, denoted as $b_{pobj}$. Then, we parse each element in the dependency structure in a bottom-up fashion until we reach the final destination.
The above grounding process can be written as
\begin{align}
b_{pobj} &= f_\theta^{h}(u_{pobj}, b_0)\label{eq:pobj}\\
b_{prep} &= f_\theta^{prep}(u_{prep}, b_{pobj})\label{eq:prep}\\
b_{h} &= f_\theta^{h}(u_h, b_{prep}) \label{eq:h}
\end{align}
where $f_\theta^{*}$ represent grounding functions with $\theta$ as parameters. $b_0$ represents the dummy belief spanning the whole map $m$. Note that Eq.~\eqref{eq:pobj} and Eq.~\eqref{eq:h} shares the same function since the preposition objective itself is also a head phrase that refers to a location. See Fig.~\ref{fig:belief_update} for an example of the dependency structure.

\subsection{Recursive Belief Update with Modifier Categorization}

In the previous section, we have formally defined the goal grounding process based on the dependency structure of destination noun phrases. In this section, we introduce a more general and flexible formulation of goal grounding. First, we notice that Eq.~\eqref{eq:pobj}-\eqref{eq:h} all share the same analytical form $b'=f_\theta^*(u, b)$ that updates a belief $b$ via a grounding function $f_\theta$ given additional information $u$. Thus, we can unify Eq.~\eqref{eq:pobj}-\eqref{eq:h} and view them as a chain of general belief updates as
\begin{equation}\label{eq:belief_update_set}
\begin{cases}
&b_k = f_\theta(u_k, b_{k-1})\\
&b_{k-1} = f_\theta(u_{k-1}, b_{k-2})\\
&\cdots\\
&b_1 = f_\theta(u_1, b_{0})
\end{cases}
\end{equation}
or alternatively,
\begin{equation}\label{eq:belief_update_chain}
    b_k=f_\theta(u_k,\cdots f_\theta(u_2, f_\theta(u_1, b_0)))\cdots)
\end{equation}
where $k$ refers to the total number of $u_{pobj}$ and $u_{prep}$ we have considered. This leads to a key point to our goal grounding algorithm: instead of directly taking the whole destination noun phrase as input, we decompose it into a sequence of grammatically associated segments $(u_k)$ and recursively apply each of them to update a notion of belief $b$. Specially, we refer to segments $u_k$ as \textit{modifiers} and refer to the believes before and post update as \textit{prior} and \textit{posterior}.

\begin{table*}[htbp]
  \centering
  \caption{Update Types}
    \begin{tabular}{lllll}
    \toprule
     Update Type $t$ & Example Modifers & Applicable Prior & Posterior Type & Update Rule \\
    \midrule
    Dummy & ``to'', ``of'' & Precise/Ambiguous & Same as prior & Identity mapping \\
    Proximity & ``near'', ``besides'', ``close to'' & Precise & Ambiguous & Locations near prior \\
    Precise & ``an working area'', ``the north meeting room'' & Ambiguous & Precise & A specific area within prior \\
    Directional & ``the north'', ``the south west'' & Precise & Ambiguous & Locations within a certain orientation \\
    \bottomrule
    \end{tabular}%
  \label{tab:mod_type}%
\end{table*}%

Although Eq.~\eqref{eq:belief_update_set} establishes a general definition of goal grounding, the construction of the update function $f_\theta$ remains challenging. This is primarily due to two factors: (a) the way modifiers update prior believes are naturally diverse in natural language, and (b) the update rules cannot be determined solely based on the modifiers' grammatical roles. The first factor suggests the use of separate update functions with different structures, while the second factor implies that NLP annotation tools are not able to directly fulfill the purpose of function selection. For example, in ``the north of the entertainment room'', ``the entertainment room'' leads to a specific location while ``the north'' refers to an orientation. Thus, the way a modifier updates its prior cannot be easily determined by its grammatical role.

\begin{figure}[ht!]
    \centering
    \includegraphics[width=\columnwidth]{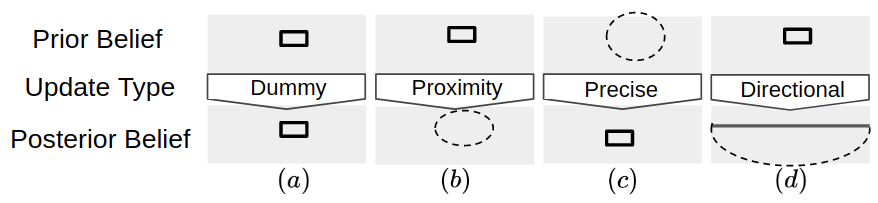}
     \caption{
     Based on the most common ways users describe destinations, we design four update functions which either (a) preserves the prior belief, (b) generate an ambiguous belief (dashed) near the prior, (c) locates a specific area (solid) within prior, or (d) generate an ambiguous belief within a certain orientation (divided by solid line) of the prior.
    }
    \label{fig:update_type}
\end{figure}

\begin{figure}[ht!]
    \centering
    \includegraphics[width=0.8\columnwidth]{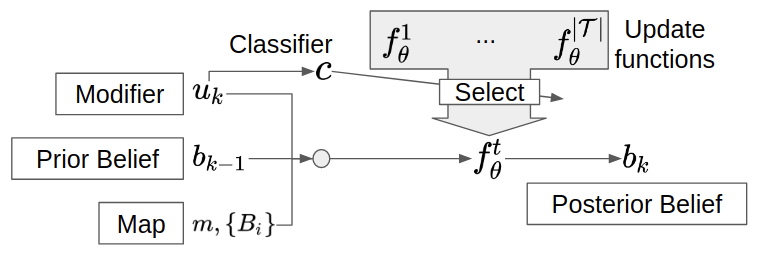}
     \caption{Single belief update step. At each step, the classifier $c$ selects one  function based on modifier $u_k$ to perform the belief update.
    }
    \label{fig:overview}
\end{figure}

To tackle the above challenge, we categorize modifiers according to how they update the prior instead of their own grammatical roles. Specifically, we classify the type of each modifier to be either ``dummy'', ``proximity'', ``precise'', or ``directional''. Each modifier type also implies constraints on the type of prior it applies to; ``precise'' believes refer to specific areas, while ``proximity'' believes refer to an orientation or proximity relation with respect to its prior. This design is supported by another key observation that in natural language instructions, the ways users implicitly describe a location is highly limited. For instance, ``the north of room 404'' is valid while ``the north of the north'' is not. See Table.~\ref{tab:mod_type} for a summary of all valid update types we use in this paper. With such setting, we are able to construct separate update functions $f_\theta^t$ for each update type $t\in\mathcal{T}$ and invoke them adaptively. More formally, we construct a set of learnable functions $\{f_\theta^t:\mathcal{U}\times\mathcal{B}\rightarrow\mathcal{B}\},\forall t\in\mathcal{T}$ and a classifier $c:\mathcal{U}\rightarrow\mathcal{T}$ that selects one update function upon observing each modifier $u_k$. Then, our update function in Eq.~\eqref{eq:belief_update_set} can be written as
\begin{equation}\label{eq:multi_path_update}
b_k=f_\theta(u_k, b_{k-1})=\sum_{t\in\mathcal{T}}\mathbb{I}(c(u_k)=t)f_\theta^t(u_k, b_{k-1})
\end{equation}
where $\mathbb{I}(\cdot)$ is the indicator function. We represent the classifier $c$ as a neural network, given by
\begin{equation}
    c(u_k) = \underset{t}{argmax}\ softmax(W_c^\top\Phi_c(u_k))
\end{equation}
where $\Phi_c$ represents the hidden state extraction by a GRU layer, and $W_c$ refers to the weights of a linear layer that maps the GRU hidden state to raw classification weights. In following sections, we describe the design of all update functions except that for dummy update, which is simply the identical mapping. Learnable functions are trained by minimizing type-specific losses through back-propagation, which will be described in detail in section \ref{sec:exp_data},  See Fig.~\ref{fig:update_type} for semantic illustration of each update function. See Fig.~\ref{fig:overview} for the compute diagram of single belief update steps. Notably, Fig.~\ref{fig:overview}(d) shows an exemplary posterior generated by modifier ``the south.''

\subsubsection{Precise Update}

This update branch handles the situation where the modifier refers to a specific area within the prior. This resembles how human locates a specific room knowing some spatial constraints. Suppose the given map $m$ is segmented into $S$ areas of interest, each represented by a tuple of descriptive strings as mentioned in section \ref{sec:prob_form}. We first convert each word in the area information to fixed length embeddings and then concatenate them. The result is a matrix representation of the map information, denoted by $\Bar{m}\in\mathbb{R}^{S\times N\times H}$. $S$ is the number of area of interest. $N$ refers to the number of tokens in area descriptors. $H$ is the dimension of word embedding. We similarly encode the modifier $u_k$ as a embedding matrix $\Bar{u}_k\in\mathbb{R}^{N\times H}$. Since precise updates pick specific areas from priors, we model this update as a $S$-level classification problem which generates a discrete distribution $w_k$ defined on all areas in the map. The calculation of for each area $a_i$ is given by
\begin{equation}\label{eq:precise_update}
    w_k(i) = \frac{1}{\eta}\gamma_k(i)\cdot \Bar{w}_k(i)\cdot w_{k-1}(i)
\end{equation}
where $\gamma_k$, $\Bar{w}_k$, and $w_{k-1}$ represents directional scaling factor, modifier-map attention, and prior weights as explained as follows. $\eta$ refers to normalization factor. The full belief $b_k$ can then be recovered by assigning $w_k(i)$ to the area on the map indicated by boundary $B_i, \forall i$ and then normalize across the whole map.

The use of $\gamma_k$ is triggered by the common use of directions as adjective of the head word (see Table.~\ref{tab:mod_type}). The formula is given by
\begin{equation}\label{eq:gamma_k}
    \gamma_k(i) = (\sigma(x_i^\top e_{\alpha_k}-\underset{x\in B_0}{min}\ x^\top e_{\alpha_k})+1)^{\kappa_k}-1+\beta_k\kappa_k+\epsilon
\end{equation}
where $\sigma$ is the sigmoid function, $x_i$ is the centroid of area $a_i$, $B_0$ is the boundary of the map, $e_{\alpha_k}$ is the unit directional vector of the predicted direction ${\alpha_k}\in[-\pi, \pi)$, ${\kappa_k}\in[0, 1]$ is a trainable variable indicating whether a direction is used in $u_k$, $\beta_k$ is a shaping factor adjusting the scaling power of $\gamma_k$, and $\epsilon$ is a positive constant. When a directional adjective is used, we label ${\kappa_k}$ as $1$, rendering $\gamma_k$ in a exponential form that weighs each areas $a_i$ according to the projection of their centroids along a predicted direction. We add $\beta_k$ as an offset to provide additional flexibility. When no directional adjective is involved, ${\kappa_k}$ is labeled as $0$ and pushes $\gamma_k(i)$ to $\epsilon$ for all $a_i$, effectively cancelling this discriminative term. Notably, all terms in Eq.~\eqref{eq:gamma_k} except $x_i$ and $b_i$ are shared by all area weights and calculated by learnable functions as
\begin{equation}\label{eq:prcs_dirparam}
\begin{cases}
\alpha_k &= tanh(W_\alpha^\top\Phi_\alpha(u_k))\cdot\pi \\
\kappa_k &= \sigma(W_\kappa^\top\Phi_\kappa(u_k)) \\
\beta_k &= exp(W_\beta^\top\Phi_\kappa(u_k))
\end{cases}
\end{equation}
where $\Phi_{*}$ represent the hidden state extracted by GRU layers and $W_{*}$ are the weights of linear layers that generate scalar outputs.

The attention term $\Bar{w}_k(i)$ in Eq.~\eqref{eq:precise_update} is given by
\begin{equation}
    \Bar{w}_k(i)=\frac{1}{\Bar{\eta}}\sum_{j, l=1}^{N, N}\mathbb{I}(\Bar{m}_{ij}^\top \bar{u}_{kl}>\lambda)
\end{equation}
where $\Bar{\eta}$ is the normalization factor, $\Bar{m}_{ij}$ refers to the $j^{th}$ word embedding assigned to area $a_i$, and $\Bar{u}_{kl}$ refers to the $l^{th}$ word embedding in modifier $\Bar{u}_k$. This term weighs each area $a_i$ by counting the matching word pairs between pre-defined area information and modifier $u_k$. Word matching is examined by filtering normalized embedding dot products by threshold $\lambda$.

Finally, the area prior $w_{k-1}(i)$ is calculate by gathering the weights of each area $a_i$ from prior belief as follows
\begin{equation}
    w_{k-1}(i)=\frac{1}{\eta_{k-1}}\sum_{u, v\in b_i}b_{k-1}(u, v)
\end{equation}
where $u, v$ refers to map coordinates and $\eta_{k-1}$ is the normalization factor.

\subsubsection{Proximity Update}
When encountering prepositions referring to proximity relation, we heuristically represent the posterior as a Gaussian distribution centered at the prior and assign a variance proportional to the prior area size. The update function is then written as
\begin{equation}\label{eq:prox}
    b_k = f_\theta^t(u_k, b_{k-1})=\mathcal{N}(x_{k-1}, diag(\rho|B_{k-1}|, \rho|B_{k-1}|))
\end{equation}
where $x_{k-1}$ and $|B_{k-1}|$ are the centroid coordinate and size of the area indicated by priot $b_{k-1}$. $\rho$ is a scaling constant.

\subsubsection{Directional Update}
Notably, besides being used as a adjective, direction words (e.g., ``north'') can also be directly used as head words, e.g., ``the north of meeting room 202.'' In this case, we encounter a directional update. We again use a heuristic Gaussian distribution to represent the prior, but with an additional mask that only preserves the belief consistent with $u_k$. See Fig.~\ref{fig:update_type} for a graphical illustration. The update function can be written as
\begin{equation}
    b_k = f_\theta^t(u_k, b_{k-1})=\mathcal{N}_{k-1}\mid cos(x-x_{k-1}, e_{\alpha_k}) > 0
\end{equation}
where $e_{\alpha_k}$ is the unit direction vector of the valid direction $\alpha_k$. $\mathcal{N}_{k-1}$ takes the same form as in Eq.~\eqref{eq:prox}. $cos(\cdot, \cdot)$ is the cosine similarity. We represent $\alpha_k$ as a learnable variable similarly to Eq.~\ref{eq:prcs_dirparam} as
\begin{equation}
    \alpha_k = tanh(W_\alpha^\top\Phi_\alpha(u_k))\cdot\pi
\end{equation}

\subsection{Composite Implication}

In previous sections, we primarily focus on implicit destination descriptions consisting of two head phrases and one preposition, e.g., ``the meeting room near the north exit.'' In practical, it is not uncommon that a longer description is used especially when one uses orientation instead of proximity to describe his/her goal, e.g., ``the meeting room to the north of the entertainment area.'' Then, we have three head phrases (i.e., ``the meeting room'', ``the north'', ``the entertainment room'') and two prepositions (i.e., ``to'', ``of''). In this case, the user is not only referring to his/her goal implicitly, but also using an implicit reference, i.e., ``the north.'' We refer to this situation as \textit{composite implication} hereafter. Note that, despite the additional complexity, such descriptions can still be directly handled by our approach, since the goal grounding process as shown in Eq.~\eqref{eq:belief_update_chain} naturally generalize to longer dependency traces.


\section{Experiments}


In this section, we demonstrate our destination grounding algorithm on a real-world navigation task in an office. The robot's goal is to ground the destination description given by a user to the office map (see Fig.~\ref{fig:intro}), and generate an area-wise navigation plan. Since our grounding algorithm recursively applies the update function as shown in Eq.~\eqref{eq:multi_path_update}, we train the learnable update functions on single update steps instead of complete navigation instructions. We proceed as follows. In section \ref{sec:exp_data}, we describe the data collection and training procedure for update functions. In section \ref{sec:exp_single}, we present benchmark results on single belief update steps. Finally, we present destination grounding results on human-provided instructions with varying complexities in section \ref{sec:exp_composite}.

\subsection{Data Collection}\label{sec:exp_data}

\begin{table}
  \centering
  \caption{Area Attributes and Modifier Dictionary}
    \begin{tabular}{ll|ll}
    \toprule
     Attribute & Examples & Modifier & Examples \\
    \midrule
    id & ``100'', ``201'' & Dummy & ``to'', ``of'' \\
    category & room, area, exit  & Proximity & ``besides'', ``near'' \\ 
    sub-category & meeting/phone (room) & Direction & ``south'', ``north east'' \\
    name & ``yosemite''  & & \\
    \bottomrule
    \end{tabular}%
  \label{tab:attr_dic}%
\end{table}%

In this section, we describe the data collection for training single belief update steps. The map we use is the floor plan of an office consisting of general working areas, meeting rooms, and designated areas such as entertainment areas. Besides reusing existing spatial structures such as rooms and designated areas, we also segment general common spaces such as corridors. The whole map is segmented into 80 areas with assigned attributes, which are later synthesized to generate realistic modifier phrases. See Table.~\ref{tab:attr_dic} for summarized area attributes.

Since the language involved in single updates are highly limited, it is possible to analytically generate training data based on a finite rule set. For composite updates, however, manual labeling is required due to less constrained grammatical structures. For each update type $t$, we randomly generate $K=10$ modifiers $u$ according to predefined dictionary (see Table.~\ref{tab:attr_dic}) and take each area $a_i$ as the \textit{key area}.

We end up with $3,200$ single update samples. The procedure to generate essential training sample for each update function is described as below.

\subsubsection{Dummy Belief Updates}
Prior and posterior believes are omitted. Each training sample is a singleton $(u_k,)$ with type label $t^*=0$.
\subsubsection{Proximity Belief Update}
The prior $b_0$ is uniformly distributed within the key area. The posterior $b_1$ is a Gaussian centered at the key area with standard deviation proportional to key area size. Each training sample is a tuple $(b_0, b_1, u_k)$ with type label $t^*=1$.
\subsubsection{Directional Belief Update}
The prior $b_0$ is uniformly distributed within the key area. A direction angle $\alpha_k$ is sampled from $uniform[-\pi, \pi)$. The posterior $b_1$ is generated by taking the Gaussian similar to that in proximity update, but with half of it masked out using a dividing line perpendicular to the direction represented by $\alpha_k$. Finally, the modifier $u_k$ is determined based on $\alpha_k$. Each training sample is a tuple $(b_0, b_1, u_k,\alpha_k)$ with type label $t^*=2$.
\subsubsection{Precise Belief Update}
We first generate the prior $b_0$ randomly as the output of proximity or directional update based on the key area. Then, we sample map locations from $b_0$ and pick the top two areas $a^1,a^2$ in which most of these sampled locations fall. We then generate modifiers $u_k$ that uniquely locates $a^1$ based on a minimal rule set (see Table.~\ref{tab:mod_rule}, brackets indicate optional usage). Additionally, we label $\kappa_k$ as $1$ if a direction word is used as adjective, or $0$ otherwise. The posterior $b_1$ is uniformly distributed within $a^1$. Each training sample is a tuple $(b_0, b_1, u_k, \alpha_k, \kappa_k)$ with type label $t^*=3$.

\begin{table}[htbp]
  \centering
  \caption{Modifier Generation Rules}
    \begin{tabular}{lll}
    \toprule
     Condition of $a^1$, $a^2$ & Modifier template & Example \\
    \midrule
    Different category & [sub] category  & ``room'' \\
    Different sub-category & [direction] [sub] category & ``meeting room'' \\
    Same sub-category & direction [sub] category & ``south west area'' \\
    \bottomrule
    \end{tabular}%
  \label{tab:mod_rule}%
\end{table}%

\subsection{Training}

We train our update function set in Eq.~\eqref{eq:multi_path_update} by minimizing the total losses of all supervised terms that are applicable for each update function type. We define the losses for all supervised terms as follows. For classifier $c$, we use cross entropy loss $L_c$ as
\begin{equation}
    L_c = -log\left(softmax(W_c^\top\Phi_c(u_k))[t^*]\right)
\end{equation}

For direction $\alpha_k$, we use a marginal $\ell_2$ loss as
\begin{equation}
    L_\alpha = max\left(||\alpha_k-\alpha_k^*||_2-\left(\frac{\pi}{8}\right)^2, 0\right)
\end{equation}

For indicator $\kappa_k$, we use cross entropy loss $L_\kappa$ as
\begin{equation}
    L_\kappa = -log\left(softmax(W_\kappa^\top\Phi_\kappa(u_k))[\kappa_k^*]\right)
\end{equation}

In training phase, we hold $10\%$ of the data as testing set and train on the remaining samples. We encode text instructions using Bert embedding \cite{devlin2018bert} separately on each word and generate word embeddings of length $H=768$. We use $8$ as GRU hidden sizes. We perform optimization using Adam with $1e-4$ learning rate for $10$ epochs. See the following section for training results.

\subsection{Single Belief Update}\label{sec:exp_single}

\begin{figure}[ht!]
    \centering
    \includegraphics[width=\columnwidth]{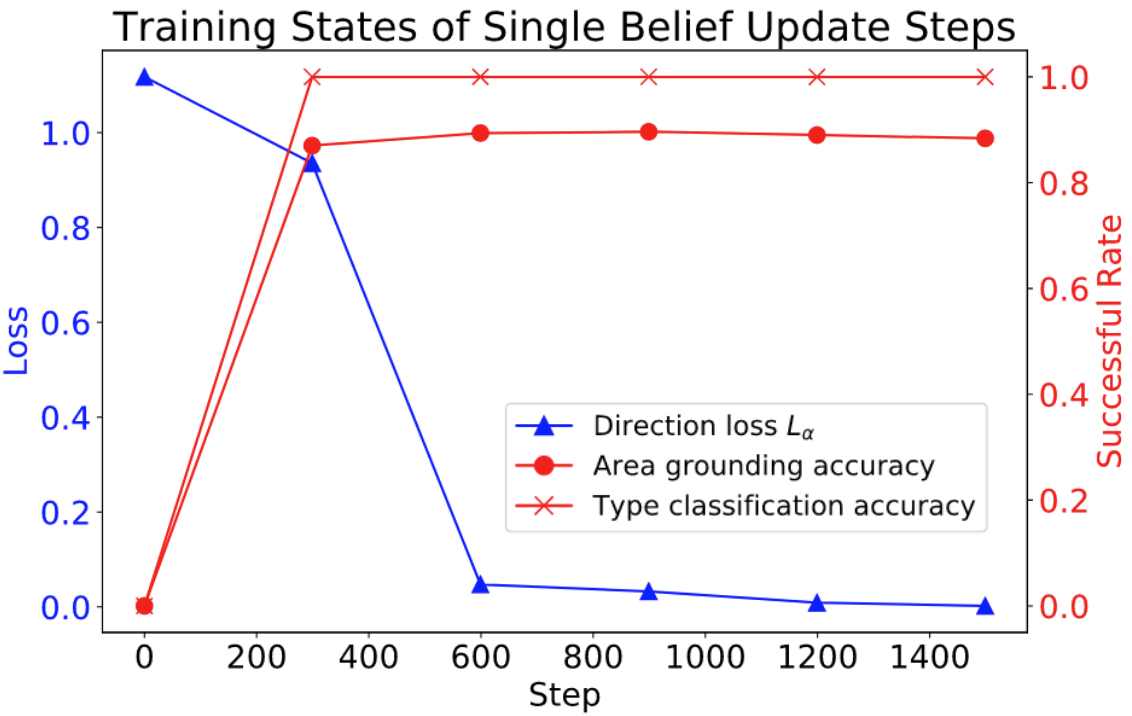}
     \caption{
     We plot the mean validation statistics of several essential grounding elements: the marginal $\ell_2$ loss on interpreted direction used in both directional and precise updates (blue), the accuracy of type classification in all update types (red crosses), and the accuracy of area grounding of precise updates (red dots). 
    }
    \label{fig:single_training}
\end{figure}

In single belief updates, the input is a prior-modifier tuple $(b_0, u)$. Each input tuple is paired with a ground true update type $t^*$ as well as the required output terms as described in section \ref{sec:exp_data}. In this section, we report the model performance on the terms that most accurately represents the quality of posterior generation. We plot the validation accuracy on test set during training on update type classification for all update types and area classification for precise update in Fig.~\ref{fig:single_training}. In the same figure, we also show the loss of direction prediction $L_\alpha$ in dual axis. Note that by using a marginal loss for $\alpha$, we accept any direction with a $22.5^\circ$ tolerance, since the direction of interest piratically spans across $180^\circ$ with $\alpha$ as the central direction.

We also provide exemplary input-output tuples from single precise update steps. In both cases, the destination category ``area'' is well registered, leading to positive belief values in all areas in category ``area''. In the first row, our precise update function locates the ``south west'' area in all areas within the prior, which is consistent with the ground truth. In the second row, however, our model fails to differentiate between potential candidates since there are multiple ``areas'' sharing the same subcategory ``working'', leading to ambiguity. This particular case is practically an artifact  of our data collection process which violates the uniqueness requirement, but indeed points out a new challenge to be solved in our future work. We omit other update types since their accuracy can be solely determined by examining single supervised terms, such as type classification and angle prediction.

\begin{figure}[ht!]
    \centering
    \includegraphics[width=\columnwidth]{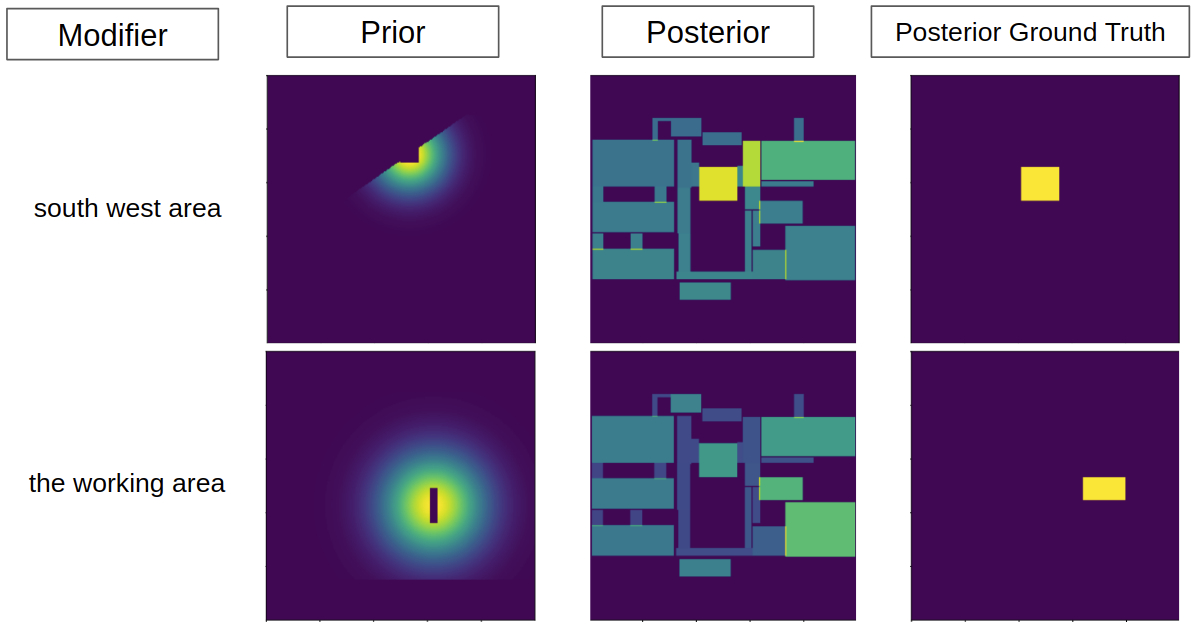}
     \caption{
     We show two examples of precise updates. Given the modifier and prior belief, our update function generate the posterior. The posterior belief is reconstructed from the area weights generated as
     Eq.~\ref{eq:precise_update} and area boundaries. Higher belief values are drawn in brighter colors.
    }
    \label{fig:single_example}
\end{figure}

\subsection{Composite Belief Update}\label{sec:exp_composite}

\begin{figure*}[ht!]
    \centering
    \includegraphics[width=170mm]{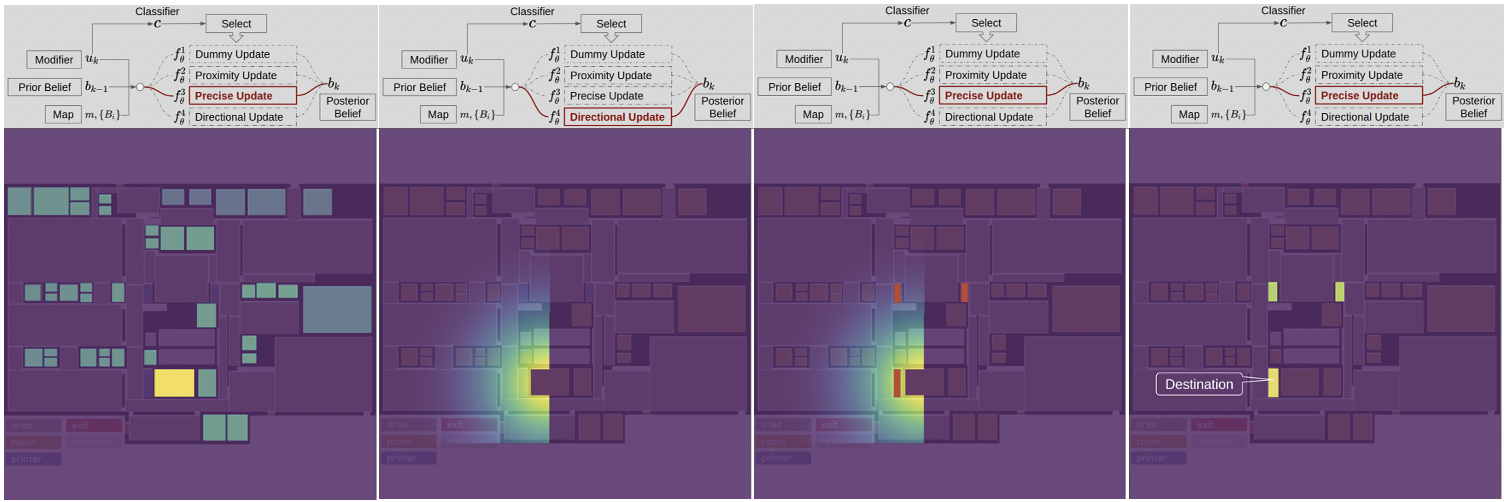}
     \caption{
     Example of posterior believes updated after each single step. Our model successfully parses the instruction description ``the printer to the west of the golden gate meeting room'' and grounds to the correct location on the map. (a), (b), and (c) show the belief after parsing ``golden gate meeting room'', ``west'', and ``printer'' respectively. Eventually the destination is located as shown in (d).
    }
    \label{fig:multi_example}
\end{figure*}

\begin{figure*}[h]
    \centering
    \includegraphics[width=170mm]{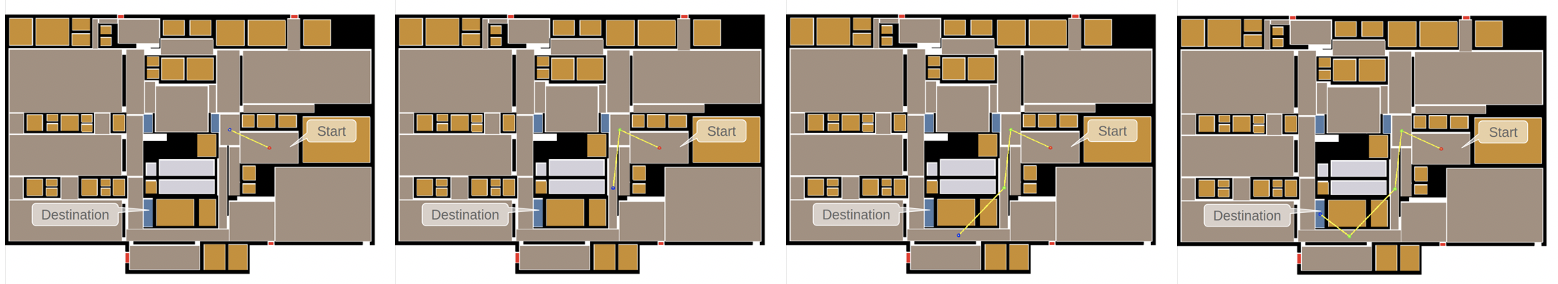}
     \caption{
     A depth-first search result of path planning.
    }
    \label{fig:plan_path}
\end{figure*}
In this section, we demonstrate a composite belief update on realistic human instructions. To collect data, we generate office layouts with a randomly chosen area highlighted. All area attributes are also given as a reference. The human subjects are asked to describe the navigation commands he/she would have used if the highlighted destination were desired. The only restriction is that the command should only describe the final destination without any information regarding the intermediate navigation. For instance, ``Go to the north phone room near the entertainment area'' is valid, while ``Go to the north exit, avoid printers'' is not. We present a navigation example in Fig.~\ref{fig:multi_example}. Given the grounded destination, we randomize a start point and invoke a depth-first search to generate the final navigation plan. Notably, the path planning module can be replaced if desired and is independent of the correctness of destination grounding. The resulted path is shown in Figure \ref{fig:plan_path}.

\begin{table}[htbp]
  \centering
  \caption{Benchmark Result on Composite Updates}
    \begin{tabular}{llll}
    \toprule
     \# Steps & \# Queries & Top1 \% & Top5 \% \\
    \midrule
    Any & 167 & 44.31 & 71.26 \\
    1 & 26 & 57.69 & 80.77 \\
    3 & 26 & 57.69 & 73.08 \\
    5 & 99 & 35.35 & 65.66 \\
    7 & 16 & 56.25 & 87.50 \\
    \bottomrule
    \end{tabular}%
  \label{tab:benchmark}%
\end{table}%

\subsection{Hardware Implementation}
\begin{figure}
    \centering
    \includegraphics[width=75mm]{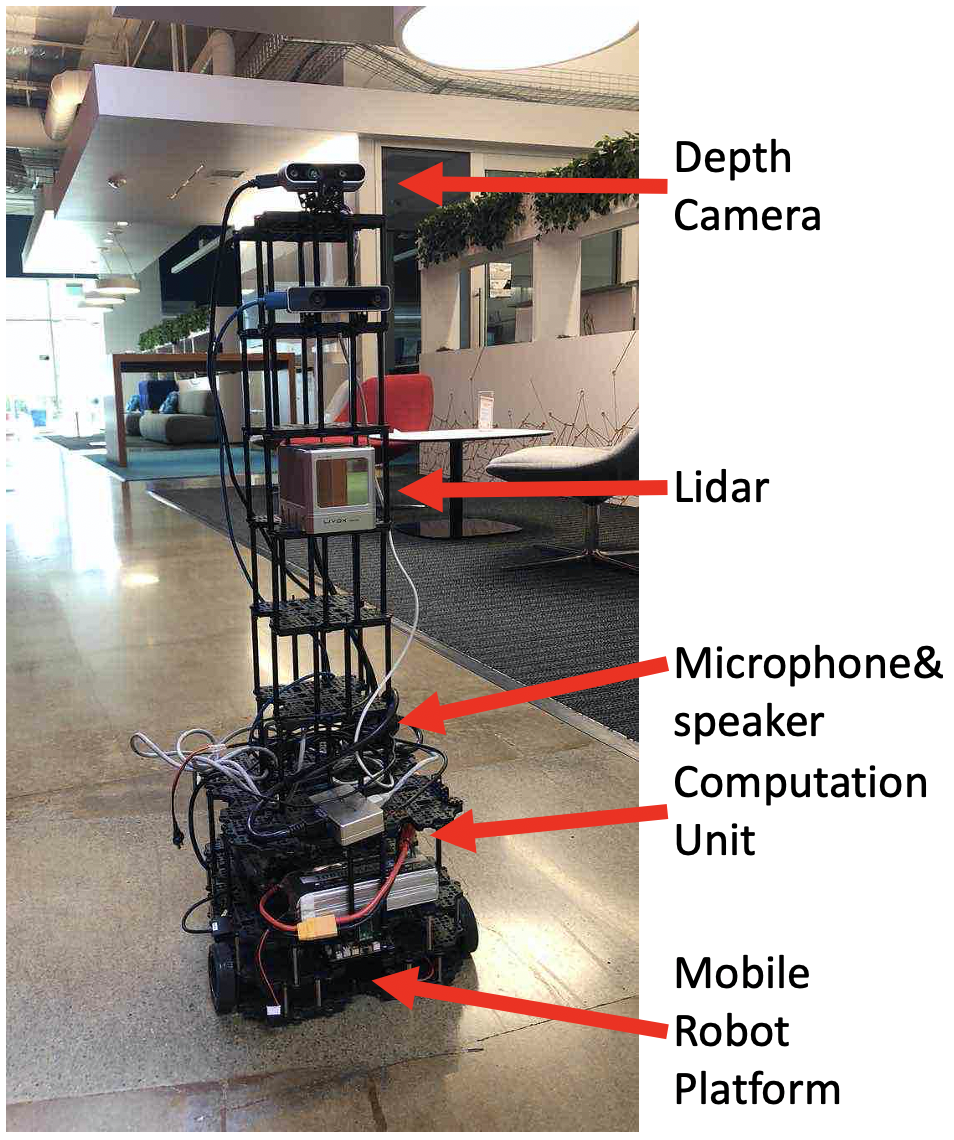}
    \caption{Mobile robot hardware platform}
    \label{fig:hardware_platform}
\end{figure}
We further integrate our language grounding model with our indoor navigation robot platform. As shown in Figure \ref{fig:hardware_platform}, the hardware platform is equipped with a depth camera and Lidar sensors for perceiving the surrounding areas. Our software algorithms are running on an Nvidia Tx2 computation unit. The calculated control commands are sent to the mobile robot for execution.


\begin{figure*}[h!]
    \centering
    \includegraphics[width=140mm]{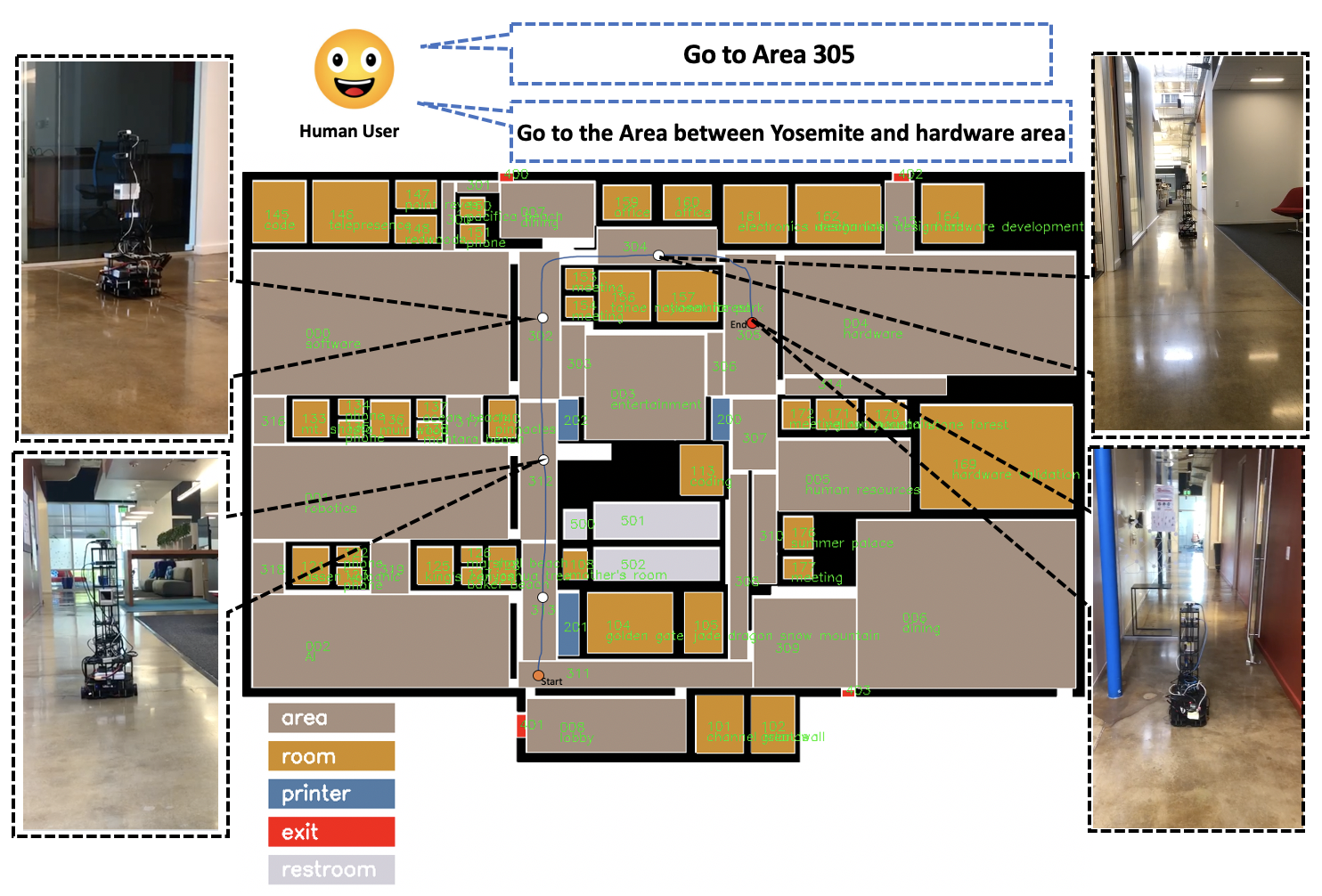}
    \caption{Given the navigation goal description, our mobile robot can plan a feasible trajectory and reach the goal.}
    \label{fig:test_path}
\end{figure*}

Our grounding model translates the human language description into a global navigation goal, sent to the stack of navigation planner for path planning. A localization algorithm based on Lidar observation provides the coordinates of the mobile robot to different modules.

In the hardware experiments, we provide two sentences describing the same navigation target. One sentence is a precise description, `` Go to the area 305''. The other sentence is a description ''Go to the area between Yosemite and hardware''. Our grounding model translates the goal description into map coordinates, and our mobile robot follows the planned trajectory to reach the target. 

The hardware test result is shown in Figure \ref{fig:test_path}, while the resulting trajectory is plotted w.r.t the 2D floor-plan. 
Therefore, the human can set the robot navigation goal without knowing the exact goal coordinates.

\section{Conclusion}
In this article, we close the gap between implicit navigation description from human natural language and robot understandable goal representation by proposing a novel framework of recursive belief updating. By exploiting the unstructured map information through instruction decomposition and modifier categorization, the proposed approach can be directly generalized with altered layout and area. 
Additionally, the proposed modularized architecture requires less effort when integrating with other modules like NLP annotation, navigation with constraints, and low-level robot motion planning.
Our hardware experiment results have shown that with the designed language model, a natural language goal description can be grounded into the navigation goal coordinates, thus making it convenient for a human user to set the navigation target of a service robot.


\bibliographystyle{IEEEtran}
\bibliography{ref}

\end{document}